\documentclass[sigconf]{acmart}
\usepackage{tabularx} 
\AtBeginDocument{%
  }

\copyrightyear{2024}
\acmYear{2024}
\setcopyright{acmlicensed}
\acmConference[CIKM '24] {Proceedings of the 33rd ACM International Conference on Information and Knowledge Management}{October 21--25, 2024}{Boise, ID, USA.}
\acmBooktitle{Proceedings of the 33rd ACM International Conference on Information and Knowledge Management (CIKM '24), October 21--25, 2024, Boise, ID, USA}
\acmISBN{979-8-4007-0436-9/24/10}
\acmDOI{10.1145/3627673.3679219}

\usepackage{titlesec}
\titlespacing{\subsection}{0pt}{0.7\topskip}{0.5\topskip}
\titlespacing{\subsubsection}{0pt}{0.4\topskip}{0.3\topskip}
\begin{document}

\title{DeliLaw: A Chinese Legal Counselling System Based on a Large Language Model}

\renewcommand{\thefootnote}{\ifcase\value{footnote}\or\textsuperscript{*}\or\textsuperscript{†}\fi}


\author{Nan Xie}
\authornote{Equal contribution}
\email{n.xie@siat.ac.cn}
\affiliation{
  \institution{University of Chinese Academy of Sciences}
  \institution{Shenzhen Institute of Advanced Technology, CAS}
  \institution{SIAT-DELI AI and Law Joint Lab}
  \country{China}
}

\author{Yuelin Bai}
\authornotemark[1]
\email{yl.bai@siat.ac.cn}
\affiliation{%
  \institution{Shenzhen Institute of Advanced Technology, CAS}
  \country{China}
}


\author{Hengyuan Gao}
\email{hy.gao1@siat.ac.cn}
\author{Ziqiang Xue}
\email{zq.xue@siat.ac.cn}
\affiliation{%
  \institution{Hebei University}
  \country{China}
}

\author{Feiteng Fang}
\email{feitengfang@mail.ustc.edu.cn}
\author{Qixuan Zhao}
\email{qixuanzhao@mail.ustc.edu.cn}
\author{Zhijian Li}
\email{lzj0@mail.ustc.edu.cn}
\affiliation{%
  \institution{University of Science and Technology of China}
  \country{China}
}

\author{Liang Zhu}
\email{l.zhu@siat.ac.cn}
\affiliation{%
\institution{South University of Science and Technology of China}
\country{China}
 }

\author{Shiwen Ni}
\authornote{Corresponding Author}
\email{sw.ni@siat.ac.cn}
\author{Min Yang}
\authornotemark[2]
\email{min.yang@siat.ac.cn}
\affiliation{%
  \institution{Shenzhen Institute of Advanced Technology, CAS}
  \institution{SIAT-DELI AI and Law Joint Lab}
  \country{China}
  }

\renewcommand{\shortauthors}{Xie, et al.}


\begin{abstract}
  Traditional legal retrieval systems designed to retrieve legal documents, statutes, precedents, and other legal information are unable to give satisfactory answers due to lack of semantic understanding of specific questions. Large Language Models (LLMs) have achieved excellent results in a variety of natural language processing tasks, which inspired us that we train a LLM in the legal domain to help legal retrieval. However, in the Chinese legal domain, due to the complexity of legal questions and the rigour of legal articles, there is no legal large model with satisfactory practical application yet. In this paper, we present DeliLaw, a Chinese legal counselling system based on a large language model. DeliLaw integrates a legal retrieval module and a case retrieval module to overcome the model hallucination. Users can consult professional legal questions, search for legal articles and relevant judgement cases, etc. on the DeliLaw system in a dialogue mode. In addition, DeliLaw supports the use of English for counseling. we provide the address of the system:
\url{https://data.delilegal.com/lawQuestion}.
\end{abstract}





\maketitle


\section{Introduction}
Traditional legal retrieval systems \cite{1} aim to retrieve legal information such as legal documents, statutes and case law. However, they are limited in their ability to semantically understand user queries.Open-source large language models, such as LLAMA \cite{2} and LLAMA2 \cite{3}, Falcon \cite{4}, Vicuna \cite{5}, MOSS \cite{6}, ChatGLM, and ChatGLM2 \cite{7}, have demonstrated satisfactory performance in general-purpose domains following pre-training on large-scale corpora. However, deploying them in very specialized fields such as medicine, legal, and finance presents challenges, primarily due to the scarcity of high-quality, fine-tuned data in these areas and the inherent problem of "hallucination" in generative models. Even the state-of-the-art GPT-4 model \cite{8} in the Chinese legal domain generates a significant number of fictitious legal texts, highlighting the prevalence of this issue.

Recent studies have explored the fine-tuning of open-source LLMs using legal data to develop legal-specific LLMs. One example is the Chinese legal model LawGPT \cite{9}, which is obtained by fine-tuning ChatGLM-6B LoRA 16-bit instructions. The fine-tuning process involves utilizing existing legal Q\&A datasets and constructing high-quality legal text Q\&A based on real legal articles and cases using ChatGPT’s API. While LawGPT incorporates more legal knowledge into ChatGLM, its performance remains suboptimal. To address this limitation, Huang et al. \cite{10} proposed a method that retrieves relevant legal articles based on the user’s query and contextual information, which can serve as evidence for the query before generating each response. The final response is then generated based on these legal articles. ChatLaw \cite{11} leverages an external knowledge base to mitigate model hallucinations. However, unlike LawGPT, ChatLaw trains a model specifically to extract legal feature words from the user’s everyday language. These extracted legal keywords are then used as queries for law retrieval.

\begin{figure}[t]
\centering
\includegraphics[width=0.47\textwidth]{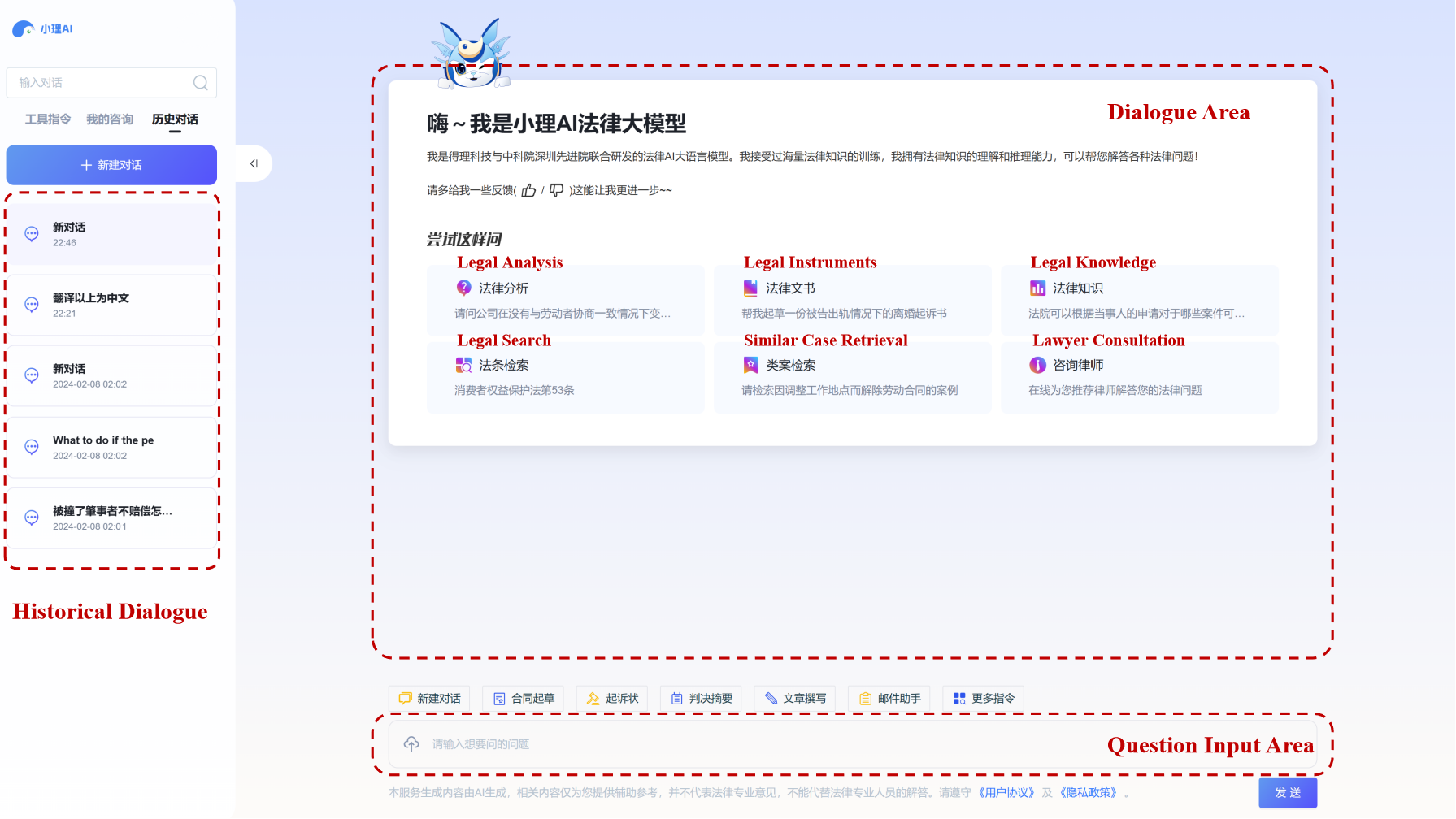} 
\vspace{-0.5em}
\caption{The interface of DeliLaw system.}
\label{fig1}
\vspace{-1.0em}
\end{figure}


While Chinese legal LLMs have made significant advancements compared to general-purpose LLMs, they still fall short of direct practical application. To address this gap, this work introduces DeliLaw, a practical legal interrogation system built upon a large language model. We first trained deep learning models dedicated to intent classification to effectively classify user queries according to different legal application scenarios. To build our premium dataset, we sourced extensive data from expert Chinese legal platforms, enriched with statutes accessed through our regulatory retrieval system, ensuring relevant legal texts were included. We crafted diverse prompts based on professional legal advice to facilitate answer generation. Lawyers were engaged to annotate the data, critical for upholding the dataset's high quality and professional reliability. For legal retrieval, we enhance the adaptability and accuracy of the system through a two-stage fine-tuning.The laws in our system response are retrieved from the law database, and since the law database is updated in real time, all the laws returned by the system are real and valid laws. Combining vector library retrieval and ElasticSearch technology, we constructed an efficient case retrieval module to provide users with comprehensive and practical legal information. In the end, by organically integrating the modules and comprehensively reasoning about DeliLaw, we successfully built an intelligent and professional system to meet users' diverse legal needs.The system’s interface is depicted in Figure 1. It enables users to engage in a free dialogue with an AI assistant to seek legal-related inquiries. The AI assistant responds professionally by providing specific legal provisions based on retrieved laws. The link to our demo video is: \url{https://youtu.be/DxuRGgLLjD0}.

\begin{figure*}[t]
  \centering
  \includegraphics[width=0.98\textwidth]{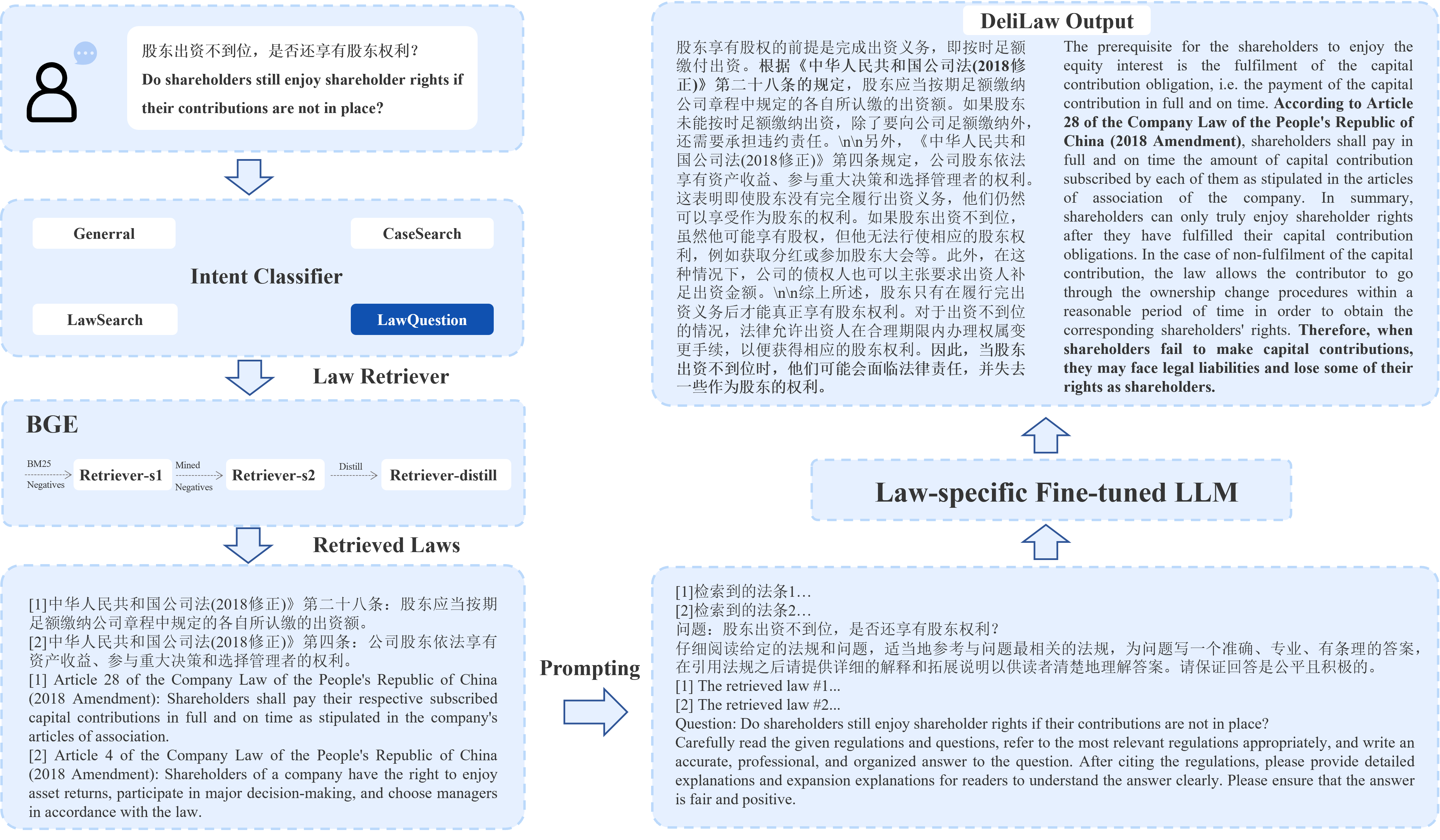}
  \caption{An illustration of DeliLaw's inference process. }
  \label{fig2}
\end{figure*}

\section{System Overview}
\textbf{Intent Classifier}
This intent classification model is based on a RoBERTa-large architecture, trained utilizing the dataset we compiled with four distinct classifications. On our internal test set, the accuracy of the model on “\textit{LawQuestion}”, “\textit{LawSearch}”, “\textit{CaseSearch}”, and “\textit{General}” the accuracy is 99.9\%, 100\%, 100\%, and 99.6\% for the four categories respectively.
\subsection{Law Retriever}
Our legal retrieval model is based on the BGE embedding model \cite{12}, currently the state-of-the-art in embedding models for dense retrieval. BGE undergoes pre-training with retroMAE \cite{13} and contrastive learning on a large corpus of Chinese texts, boasting formidable retrieval capabilities for Chinese. We fine-tune the BGE-large-zh-v1.5 model using the infoNCE loss function to adapt it to the legal domain retrieval. The fine-tuning of the BGE model utilizes collected query-statute pairs from two main sources: 1. Question and answer data from the internet, where the queries are usually shorter and the scope of the labeled statutes is smaller. 2. Queries generated using LLM-based data augmentation methods (e.g., Promptagator \cite{14}, UDAPDR \cite{15}), which are longer and can use a diverse set of statutes as labels. In the LLM-based data augmentation approach, we initially prompt GPT-4 to generate queries for a small set of statutes, then use these generated queries to fine-tune a smaller LLM in a LoRA manner. The fine-tuned smaller LLM is then used to generate queries for fine-tuning the retrieval model, achieving a balance between quality and economy.
In our training setup, fine-tuning is divided into two stages. In the first stage, the positive samples are the query’s labeled statutes, and negative samples use in-batch negative statutes, along with hard negative statutes mined using BM25. The second stage mines new hard negative statutes using the retrieval model fine-tuned in the first stage, with other settings being the same, to re-fine-tune the retrieval model.
During retrieval, both the query and the statute are input into the fine-tuned model, from which the embedding of the last layer's [CLS] token is extracted, yielding a vector of length 1024. The similarity between the query and statute is calculated using the cosine similarity of the vectors.  The statute embeddings are stored in the vector database Milvus \cite{16}, and for each query, the top three statutes with the highest similarity are retrieved from the database.
The final experimental results, evaluated under three regulatory frameworks, demonstrate an MRR of 61.6\% and a RECALL of 71.1\%. These figures highlight the effectiveness of our approach in enhancing the precision and recall of legal information retrieval, thereby underscoring the model's capability to accurately identify relevant statutes in response to user queries.
\subsection{Case Retriever}
Due to the large size of our case database, which consists of hundreds of millions of lengthy cases, direct retrieval using the Milvus vector library for similar laws and regulations is not efficient. To overcome this challenge, we adopt a two-step retrieval approach. Firstly, we extract keywords from the query. Then, we utilize the ElasticSearch library to retrieve relevant cases based on these keywords.

\subsection{Law-specific Fine-tuned LLM}
To construct our legal fine-tuning data, we begin by leveraging ChatGPT’s API to generate Q\&A data and multi-round dialogue data using real laws as the basis. This initial data are then carefully reviewed and modified by professional lawyers to ensure its accuracy and relevance. Finally we fine-tune the ChatGLM2-6b model using the constructed legal data together with other publicly available generic data. 

\subsection{Inference Process}
Figure 2 shows an inference process of DeliLaw system. Specifically, the user's question input will go through an intentional classification model, and the user's question will be classified into four categories: “\textit{LawQuestion}”, “\textit{LawSearch}”,“\textit{CaseSearch}”, and “\textit{General}”. If the predicted category is “\textit{General}”, it means that the user is asking a general question, e.g., “\textit{Hi, what's your name?}”, which will be fed directly to our fine-tuned Legal Large Language Model for answering. If the category is “\textit{CaseSearch}”, e.g., “\textit{Please give me cases related to hit-and-run.}”, then the user's question is used as a query to call the Case Retriever to retrieve legal cases and return them to the user directly. If the category is “\textit{LawSearch}”, e.g., “\textit{Article 3 of the Civil Code of the People's Republic of China...}”, then the user's question is used as a query to call the Law Retriever to retrieve legal articles and return them to the user directly. If the category is "\textit{LawQuestion}" as in Figure 2, the user's question is used as a query, and then the Law Retriever is called, which combines the retrieved law and the user's question using the prompting process and feeds it to the fine-tuned Legal LLM, which will then refer to the law and give a final response.

\section{DeliLaw Datasets}
\subsection{Open General Data}
To maintain the LLM's general capabilities while boosting its proficiency in Chinese dialogue, we've collected extensive Chinese public datasets for training. Utilizing these datasets holistically aims to render the LLM versatile and resilient in handling Chinese conversations.

\subsection{Legal Data}
Some work \cite{17} \cite{18} has shown that when small samples of high-quality data are incorporated into the fine-tuning process, the performance and generalisation of the model can be significantly improved. We focus on constructing small samples of high-quality legal data, which we introduce during the fine-tuning phase.

\subsubsection{Legal Q\&A data}
We collected a large amount of legal Q\&A data on legal websites. After cleaning the data with duplicated answers and low quality answers, we extract the legal texts cited in the answers by rules and call our statute retrieval interface to get the complete content. Finally, we input the questions and related statutes into GPT-3.5 to expand the answer content and get better answers.

However, the questions asked in the collected data are relatively short, which does not improve the model's ability to understand longer questions.

For this reason, we extracted 28k pieces of initially cleaned data and extended our problem using deep evolution and breadth evolution \cite{19}. The questioning is made complex and diverse by deep evolution and breadth evolution. The evolved data are filtered to remove data with no information gain compared to the original problem.

\subsubsection{Human annotated data}
We manually annotated 3,492 data involving real-world legal scenarios quizzes and added these to supervised fine-tuning to optimise the model. To ensure the professionalism and accuracy of the data, we invited legal experts to perform the annotations. These data cover seven common areas of law: intellectual property, criminal offences, labour disputes, property disputes, corporate compliance, matrimonial and family matters, and urban renewal.

\subsubsection{Cause of actions in GPT-4}
We strategized to augment the model's legal acumen by curating data across diverse legal contexts. Initially, we amassed 602 criminal and 200 administrative case inquiries. To bolster the reliability of generated answers, we interfaced with legal databases, extracting the five most pertinent statutes per question. These statutes served as a foundation for GPT-4 to craft informed responses. To refine these outputs, expert legal professionals vetted and enhanced the GPT-4-generated content, ensuring high-quality, precise, and current regulatory insights.

\subsubsection{Judicial syllogism Data}
LLMs excel in parsing brief queries but falter with lengthy ones, often due to gaps in specialized legal knowledge, leading to less than optimal responses. To address this, we collated complex legal inquiries from various sources including legal news, social media, and industry forums, covering 263 typical civil litigation issues. Legal experts crafted detailed answers for these questions, employing the judicial syllogism framework: statutes as major premises, case facts as minor premises, and judgements as conclusions. By adhering to this logical structure, we ensured that the generated answers were precise, coherent, and aligned with legal reasoning standards.
\subsubsection{Guided conversations}
In legal practice, client-attorney dialogue is central, where clients often present issues tersely, lacking depth. Lawyers must probe beneath surface questions, skillfully drawing out detailed, pertinent facts. Guided discussion uncovers case nuances, enhancing understanding and pinpointing legal challenges for tailored advice.
Hence, in formulating the dataset, we drew inspiration from the multi-turn dialogue between Doctors and patients in HuatuoGPT \cite{20}. Building upon this foundation, we devised roles for clients and attorneys utilizing GPT-4 to generate guiding dialogues within the context of legal consultations.

\section{Evaluation}
\begin{figure}[t]
\centering
\includegraphics[width=0.47\textwidth]{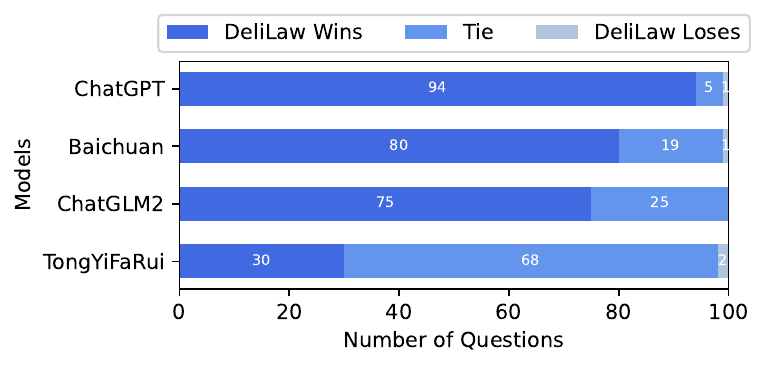} 
\caption{Human assessment, comparing DeliLaw to 4 different baselines across 100 test questions.}
\label{fig 3}
\end{figure}

We selected 100 questions for a comprehensive review of five LLMs, including DeliLaw, and analysed them in depth on the basis of the following explicit scoring criteria.
\begin{itemize}
\item 3 : Answer is fully relevant to the search, both the cited statute and law are highly relevant, and the logic is applied correctly.
\item 2 : Answer is generally relevant to the content of the search, the citation of the statute is correct but the law is incorrect and the logic is correctly applied.
\item 1 : Answer is relevant to the content of the search, the citation of the statute is incorrect or the law is incorrect or the logic is incorrect.
\item 0 : Answer is not relevant to the content of the search, the statute is incorrectly cited or the statute is incorrect or the logic is incorrect.
\end{itemize}

Figure 3 shows the results of comparing DeliLaw with the other four models in terms of scoring on the manual evaluation. It can be seen that compared to ChatGPT, Baichuan, ChatGLM2, and DeliLaw, DeliLaw performs better in answering the questions with accurate references to regulations and laws, and logical applicability in a significant number of cases.DeliLaw produced better responses than TongYiFaRui in 30 evaluation questions.
From the overall evaluation, it can be seen that DeliLaw's responses to legal advice questions are preferable to the other models.

\section{Conclusion}
DeliLaw, a Chinese legal counseling system, facilitates easy searching for legal articles, case retrieval, and access to professional legal advice. It efficiently connects users with a vast array of legal resources and expert opinions, ensuring that both seasoned professionals and the public can navigate the complexities of the law with ease. The system currently boasts more than 20,000 professional users and receives nearly 3,000 calls daily. Future research will concentrate on enhancing the system's contextual understanding, broadening the coverage of legal fields to encompass more legal knowledge, and further optimizing the user experience.

\begin{acks}
This work was supported by National Key Research and Development Program of China (2022YFF0902100), China Postdoctoral Science Foundation (2024M753398), Postdoctoral Fellowship Program of CPSF (GZC20232873), GuangDong Basic and Applied Basic Research Foundation (2023A1515110718, 2024A1515012003 and 2024A1515030166), National Natural Science Foundation of China (62376262), Shenzhen Basic Research Foundation (JCYJ20210324115
614039). 
\end{acks}
\newpage
\bibliographystyle{ACM-Reference-Format}
\bibliography{CIKM2024}


\end{document}